\documentclass[twoside,leqno,twocolumn]{article}

\usepackage[letterpaper]{geometry}
\usepackage{ltexpprt}

\usepackage{cite}
\usepackage{amsmath,amssymb,amsfonts}
\usepackage{graphicx}
\usepackage{textcomp}
\usepackage{xcolor}
\usepackage{algorithm,algcompatible,amsmath}

\usepackage{mathtools}
\usepackage{thmtools} 
\usepackage{booktabs}

\def\BibTeX{{\rm B\kern-.05em{\sc i\kern-.025em b}\kern-.08em
    T\kern-.1667em\lower.7ex\hbox{E}\kern-.125emX}}
\declaretheorem{definition} 

\begin{document}

\title{\Large Semantic Discord: Finding Unusual Local Patterns for Time Series}
\author{Li Zhang\thanks{
Department of Computer Science, George Mason University, \{lzhang18, ygao12, jessica\}@gmu.edu}\qquad Yifeng Gao\footnotemark[1]\qquad Jessica Lin\footnotemark[1]}
\date{}

\maketitle


\fancyfoot[R]{\scriptsize{Copyright \textcopyright\ 2020 by SIAM \\
Unauthorized reproduction of this article is prohibited}}





\begin{abstract} \small\baselineskip=9pt
Finding anomalous subsequence in a long time series is a very important but difficult problem. Existing state-of-the-art methods have been focusing on searching for the subsequence that is the most dissimilar to the rest of the subsequences; however, they do not take into account the background patterns that contain the anomalous candidates. As a result, such approaches are likely to miss local anomalies. We introduce a new definition named \textit{semantic discord}, which incorporates the context information from larger subsequences containing the anomaly candidates. We propose an efficient algorithm with a derived lower bound that is up to 3 orders of magnitude faster than the brute force algorithm in real world data. We demonstrate that our method significantly outperforms the state-of-the-art methods in locating anomalies by extensive experiments. We further explain the interpretability of semantic discord.
\end{abstract}

\section{INTRODUCTION}

Time series anomalous sequence detection is an important problem and has wide application in different domains such as medical care \cite{hong2015multivariate}, fraud detection \cite{ghosh1994credit, chandola2009anomaly} and Internet of Things (IoT) \cite{munir2017pattern}, as the anomalies identify unexpected and unusual items or events that are different from normal patterns. Particularly, the task involves identifying a time series subsequence that is the least similar to all other subsequences in a long time series. Such anomalous subsequence has also been referred as \textit{discord}, which is defined as the subsequence of a given length that has the largest z-normalized Euclidean Distance to its closest match \cite{keogh2005hot}. The length of the subsequence is determined by some prior knowledge on the time series. 
\begin{figure}[h]
    \centering
    \includegraphics[width=80mm]{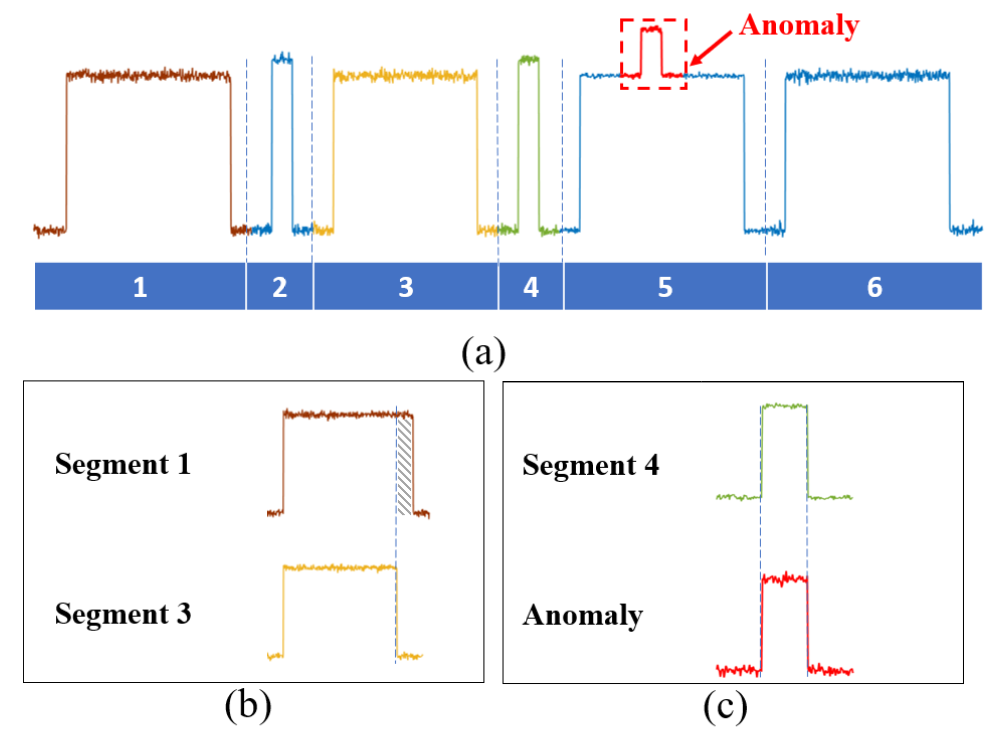}
    \caption{An illustration of a local anomalous subsequence in time series. \textbf{(a)} Time series with labeled segments 1 to 6. \textbf{(b)} Segments 1 and 3 after normalization have large distance. \textbf{(c)} Segment 4 and local anomaly have close to zero distance after normalization.}
    \label{fig:intro_context}
    \vspace{-2.5mm}
\end{figure}

Unfortunately, the approaches following the current discord definition, in which a subsequence is normalized using all the points in the subsequence, would highly likely miss true anomalies in at least some cases. We demonstrate the problem by a simple example generated from a cycle in dishwasher power consumption. Consider a time series shown in Figure \ref{fig:intro_context} (a) with six segments labeled 1 to 6 respectively. 
Segments 1, 5 and 6 have the same shape, except that Segment 5 contains a small bump, which is a local anomaly. Segment 3 has a slightly different length from Segments 1 and 6. The local anomaly in Segment 5 has identical shape as Segments 2 and 4 after z-normalization. 

Clearly, it is impossible to find the local anomaly with its true length via the classic discord definition, as the local anomaly and Segment 4 (or Segment 2) would be indistinguishable and resulting in a distance of zero after z-normalization. (Figure 1(c)). Moreover, if we expand the input length to some length close to the size of Segment 5, Segment 3 would be identified as the anomalous subsequence because of slight length difference to it closest match, Segment 1, as shown in Figure \ref{fig:intro_context}(b). Furthermore, even if some longer subsequence containing the bump could be identified given some arbitrary length, the length used to find the anomaly would be so far off from the ground truth that it loses interpretability.


To resolve this issue, in this paper, we propose a new anomaly definition, \textbf{semantic discord}. Our new discord definition incorporates the background subsequence containing the anomaly candidate, namely, a context subsequence, which provides local semantic information for detecting anomalous subsequence. Under the new definition, we search over all possible context subsequences and use the best ones to normalize the anomalous candidates respectively (as opposed to normalizing using the candidates themselves).


One may consider finding context a trivial problem, but it is not. Bringing in context subsequences forces the search space to grow quadratic in the length of context subsequence. This is because for one anomaly subsequence candidate of length $l$, if the context subsequence is length $L$ ($L>l$), there are $L-l+1$ possible contexts to choose from. To compare two candidates, we need to search over $O((L-l+1)^2)$ possible pairs of context subsequences. To address this issue, in this paper, we derive an effective novel lower bound to reduce the extra cost for computing the quadratic number of distances. 

The contributions of this paper are summarized as follows:
\vspace{-1mm}
\begin{itemize}
    \item We introduce a new definition named \textit{semantic discord}, which incorporates the context information from larger subsequences containing the anomaly candidates.
    \vspace{-1mm}
    \item We propose an efficient exact algorithm that is up to 3 orders of magnitude faster than the brute force algorithm in real world data. 
    \item We demonstrate that our method significantly outperforms the state-of-the-art methods in locating anomalies by extensive experiments. We further explain the interpretability of semantic discord. 
\end{itemize}
The paper is organized as follows. In Section 2, we review the related work and provide the background information. In Section 3, we provide the definitions on time series and formally define our problem. In Section 4, we provide details on our problem using smart brute force, followed by the derivation of an effective lower bound, as well as the proposed algorithm. In Section 5, we perform extensive experiments to show that our method outperforms the state-of-the-art methods. We also test the scalability with synthetic and real datasets. 
\section{BACKGROUND AND RELATED WORK}
We describe the state-of-the-art methods in time series anomaly detection in this section. HOTSAX~\cite{keogh2005hot} is one of the first algorithms to investigate on the detection of anomalous time series subsequence. The authors defined the \textit{discord} to be the subsequence that has the largest 1-nearest-neighbor distance in a single long series, and proposed an efficient algorithm by reordering the candidate subsequences. Recently, Matrix Profile based techniques \cite{zhumatrix,yeh2016matrix} have provided acceleration on computing the 1-nearest-neighbor distances. The results can be used to accelerate computation of HOTSAX.  Lin et al.~\cite{lin2008exact} defined anomalies as subsequences with zero or a small number of reverse nearest neighbors. Senin et al.~\cite{senin2015time} exploited grammar induction to generate rule density curve which can be used to identify anomalous subsequences corresponding to rare grammar rules. None of the existing works mentioned above considered any context information; in fact, all the existing definitions of time series anomalies would favor global anomalies and miss local unusual patterns. 

There is extensive work on finding local, point-based outliers such as \cite{breunig2000lof, kriegel2008angle,kriegel2009outlier, keller2012hics, chandola2009anomaly} based on local density or looking for a subspace in high dimension. Unfortunately, representing time series as high dimensional points will not solve the problem as time series subsequence anomalies are not in the sparsest region\cite{keogh2005hot}. These methods are not designed for high-dimensional overlapping subsequences. 

Several other existing works~\cite{wang2016self,gao2017trajviz,feremans2019pattern} on finding anomalous subsequences are worth mentioning, though the problem settings in their work are not exactly the same as ours. Wang et al.~\cite{wang2016self} developed an anomaly detection algorithm on aligned time series instances on manufacturing data. Fereman et al.\cite{feremans2019pattern} developed features on discrete time log covariates to guide an isolation forest on anomalous sequence discovery. In addition, with the rising interest in deep learning, many deep-learning-based unsupervised anomaly detection algorithms ~\cite{malhotra2016lstm, munir2018deepant, yuan2018muvan, chalapathy2019deep} have been proposed. However, training such neural networks is very time-consuming, and the results lack direct interpretability comparing with traditional search-based methods~\cite{feremans2019pattern}. Since our method does not reply on alignments or covariates, and it has the desirable property of interpretablity, we do not compare with the above methods.  

\section{DEFINITIONS}
We begin with the fundamental definitions of time series. \vspace{-1mm}
\begin{definition}
A \textbf{Time series} $\mathcal{T} = [t_1, t_2, \ldots, t_n]$ is an ordered list of data points, where $t_i$ is a finite real number and $n$ is the length of time series $\mathcal{T}$.
\end{definition}

\begin{definition}
A \textbf{time series subsequence} $S_i^L= [t_i, t_{i+1}, \ldots, t_{i+L-1}]$ is a contiguous set of points in time series $\mathcal{T}$ starting from position $i$ with length $L$. Typically $L\ll n$, and $1\leq i\leq n-L+1$.
\end{definition}

Subsequences can be extracted from time series $\mathcal{T}$ by sliding a fixed-length window through the time series. 

Given two subsequences of the same length, the Euclidean Distance is often used to measure their differences. To achieve scale and offset invariance, each subsequence must be properly normalized before the actual distance computation. The normalization step in general is very critical, as noted in previous work --- ``\textit{without normalization time series similarity has essentially no meaning. More concretely, very small changes in offset rapidly dwarf any information about the shape of the two time series in question.}"~\cite{keogh2003need}

For discord computation, previous work~\cite{keogh2005hot} uses z-normalized Euclidean Distance in order to make scale-invariant subsequences comparison prior to the distance computation. We describe the z-normalized Euclidean Distance as follows:
\vspace{-1mm}
\begin{definition}
A \textbf{z-normalized Euclidean Distance} $d_{ED}(p, q)$ of subsequences $S_p^l$, $S_q^l$ of length $l$ is computed as 
\vspace{-0.5mm}
$$\tiny {\sqrt{\sum_{m = 1}^{l}{(\frac{t_{p+m-1} - \mu_p}{\sigma_p} - \frac{t_{q+m-1} - \mu_q}{\sigma_q} })^2},}$$
\vspace{-0.5mm}
where $\mu_p$, $\sigma_p$ and $\mu_q$, $\sigma_q$ are the means and standard deviations of subsequences $S_p^l$ and $S_q^l$ respectively. 
\end{definition}



Under the current definition of z-normalization, the local anomalous subsequence would be normalized by its own mean and standard deviation. As a result, the local anomalies may be considered similar to some very different patterns. As shown in Figure \ref{fig:intro_context}, the bump of Segment 5 can be matched to Segment 2 or Segment 4, and will not be identified as discord. To overcome this problem, we propose a new definition of discord under a novel distance named  \textit{Optimal Context-Aware Distance}, to measure the dissimilarity of subsequences with contextual information. Instead of normalizing by a subsequence itself, we normalize it by using a longer context subsequence containing it, as the context subsequence would reflect the background information about the anomalous subsequence candidate. An ideal candidate of context subsequence can be in the length of some existing patterns, for example, Segment 5 in our previous example. 

To be precise, we introduce the definitions related to our proposed Context-Aware Distance as follows.
\vspace{-1mm}
\begin{definition}
Given two subsequences $S_{p}^l,  S_{q}^l$ of length $l$ and two context subsequences $S_i^L,  S_j^L$ of length $L$, where $p-L+l\leq i \leq p$  and  $q-L+l\leq l \leq q$.
The \textbf{Context-Aware Euclidean Distance} $d_{i,j}(p, q)$ between $S_p^l$ and $S_q^l$ under their contexts $S_i^L$ and $S_j^L$ is defined by 
\vspace{-0.5mm}
$$\tiny\sqrt{\sum_{m = 1}^{l}{(\frac{t_{p+m-1} - \mu_i}{\sigma_i} - \frac{t_{q+m-1} - \mu_j}{\sigma_j} })^2},$$ where $\mu_i$, $\sigma_i$ and $\mu_j$, $\sigma_j$ are means and standard deviations of subsequences $S_i^L$ and $S_j^L$ respectively. 
\end{definition}
To differentiate, the candidate subsequence $S_p^l$ , $S_q^l$ are called \textbf{target} and \textbf{reference target} respectively. Their context subsequences $S_{i}^L$ and $S_j^L$ are called the \textbf{context} and \textbf{reference context} respectively. 

To simplify the notation, we denote a subsequence by $T$ if it is a (reference) target and a subsequence by $C$ if it is a (reference) context. We omit the length for the brevity of computation that we will describe later. For example, a target $S_p^l$ is denoted as $T_p$ and a context $S_i^L$ is denoted as $C_i$. 

As previous work \cite{chiu2003probabilistic,kitaguchi2004extracting, keogh2005hot} noted, finding discord requires excluding self-matches, which refer to the subsequence itself or those that overlap with the subsequence. In our work, we only consider non-self matches of contexts and targets. We formally define non-self match as follows:
\vspace{-0.7mm}
\begin{definition}
Given a time series $\mathcal{T}$, a subsequence $S_i$ with length $L$ is considered a \textbf{non-self match} of another subsequence $S_j$ of length $L$ if $|i-j|> L$.
\end{definition}
Clearly, if $C_j$ is a non-self match of $C_i$, then $T_q$ would be a non-self match of $T_p$. In other words, if two contexts are not overlapping, any targets within the context pair respectively will not be overlapping. Thus, we only need to enforce non-self match between context $C_i$ and $C_j$ to guarantee non-self matches between any pairs of their targets, respectively.
\vspace{-1mm}
\begin{definition}
Given two non-overlapping (non-self match) targets $T_{p}$ and $T_q$ of length $l$ and a context length $L$, the \textbf{Optimal Context-Aware Euclidean Distance} between $T_p$ and $T_q$ is defined as 

$$\small d_{opt}(p, q) = \min\limits_{(i, j)\in \Omega_{p,q}} d_{i,j}(p,q), $$
where $\Omega_{p,q} = \{(i,j)| (p-L+l\leq i\leq p)  \land (q-L+l\leq j \leq q) \land d_{ED}(i, j) < \epsilon\}$.
\end{definition}

The intuition behind $\Omega_{p,q}$ is simple: given targets $T_p$, $T_q$, we search over all possible combinations of context pairs that are at least similar within some large constant threshold $\epsilon$ and overlap with targets $T_p$ and $T_q$, respectively.

We are ready to define our proposed definition of semantic discord:
\begin{definition}
Given a time series $\mathcal{T}$, the target $T_{p}$ of length $l$ is said to be the \textbf{semantic discord} of $\mathcal{T}$ if $T_{p}$ has the largest Context-Aware Euclidean Distance to its closest non-self match reference target. 
\end{definition}

By using the Context-Aware Euclidean Distance instead of the z-normalized Euclidean Distance, the semantic discord incorporates the background information from context subsequences containing the anomaly candidates, and hence better captures local anomalies in normal patterns. 
\vspace{-2mm}
\section{PROPOSED METHOD}
In this section, we start with a smart brute force algorithm under our new semantic discord definition. We then derive our proposed lower bound on the Optimal Context-Aware Distance. Finally, we introduce our proposed algorithm for finding Semantic Discord.
\begin{algorithm}[t]
    \caption{Brute Force Algorithm for Computing $d_{opt}(p,q)$}
  \begin{algorithmic}[1]
    \STATE \textbf{Input}: Time Series $\mathcal{T}$, Context length $L$, Target length $l$, $\Omega_{p,q}$
    \STATE $best\_so\_far=0$
    \FOR{$p = 1$ to $|\mathcal{T}| - l +1$} 
    \STATE    $nn\_dist$ = $Inf$

        {\color{blue} /* Compute minimum $d_{opt}(p,q)$ for each subsequence $T_q$ */}
        
    \FOR{$q = 1$ to $|\mathcal{T}| - l +1$}

     \IF{\textit{IsSelfMatch(p,q)}}  
     \STATE \textbf{continue}\quad {\color{blue} /* Skip Self-Matching */}
     \ENDIF
     
     {\color{blue} /* Compute $d_{i,j}(p,q)$ */}
     \FOR{$i = p-L+l$ to $p$} 
     \FOR{$j = q-L+l$ to $q$}

     \STATE
     $d_{i,j}(p,q)=GetDist_{CA}$($p$, $q$, $i$, $j$, $\Omega_{p,q})$
     \ENDFOR
     \ENDFOR
\STATE $dist = \min_{i,j} d_{i,j}(p,q)$
     \IF{$dist < nn\_dist$}
     \STATE $nn\_dist=dist$
    \STATE $nn\_match\_T=(p,q)$ 
     \STATE $nn\_match\_C=(i,j)$
     \ENDIF
     \ENDFOR
     \ENDFOR
     \IF{ $nn\_dist > best\_so\_far$}
        \STATE $best\_so\_far = nn\_dist$
        \STATE $best\_match\_T = nn\_match\_T$
     \STATE $best\_match\_C = nn\_match\_C$
     \ENDIF

     \STATE \textbf{return} $best\_so\_far$, $best\_match\_T$, $best\_match\_C$
  \end{algorithmic}
\end{algorithm}
\vspace{-3mm}
\subsection{Smart Brute Force}
A brute force algorithm for identifying semantic discord is incredibly costly. For every pair of target subsequences $T_p, T_q$ of length $l$, there are $O(L^2)$ possible combination of context $C_i$ and $C_j$ of length $L$. By computing all pairwise distances using the Optimal Context-Aware Euclidean Distance, the complexity of the brute force algorithm would reach $O(n^2L^2l)$. This is intractable for even a time series of a few thousand points. The algorithm is shown in Algorithm 1. 

It is possible to use the technique introduced in \cite{rakthanmanon2012searching ,yeh2016matrix} to come up with a smart brute force algorithm to speed up the distance computation. 
By expressing the Context-Aware Distance $d_{i,j}(p,q)$ between $T_p$ and $T_q$ as a function of means, standard deviations and inner products, $d_{i,j}(p,q)$ can be computed via:
\vspace{-3mm}

\small\begin{equation}
\label{eqn:exact d}
\scalebox{0.85}{$
\begin{split}
d_{i,j}^2(p,q)&=\frac{l}{\sigma_i^2}(\sigma_p^2+(\mu_p-\mu_i)^2)\\
&-\frac{2l}{\sigma_i\sigma_j}(\frac{QT_{p,q}}{l}-\mu_i\mu_q-\mu_{j}\mu_{p}+\mu_{i}\mu_{j})\\
&+\frac{l}{\sigma_j^2}(\sigma_q^2+(\mu_q-\mu_j)^2),
\end{split}
$}
\vspace{-2mm}
\end{equation}
where $\mu_i$, $\sigma_i$ and $\mu_j$, $\sigma_j$ are the means and standard deviations of $C_i$ and $C_j$ respectively. Similarly, $\mu_p$, $\sigma_p$ and $\mu_q$, $\sigma_q$ are the means and standard deviations of $T_q$ and $T_p$ respectively. $QT_{p,q} = \sum_{m=1}^l t_{p+m-1}t_{q+m-1}$ is the inner product between target $T_p$ and $T_q$, where $l$ is the length of target. We can easily pre-compute the means and standard deviations for all subsequences in $O(n)$ time, thus the major cost would be computing the inner product $QT_{p,q}$. Note $QT_{p,q}$ can be computed based on previous $QT_{p-1,q-1}$ in constant complexity~\cite{zhumatrix, yeh2016matrix}: 
$$QT_{p,q} = QT_{p-1,q-1}-t_{p-1}t_{q-1}+t_{p+l-1}t_{q+l-1}.$$ Therefore, the cost to compute $d_{i,j}(p,q)$ is constant and the overall complexity of the brute force algorithm is reduced to $O(n^2L^2)$. Although it is faster than the original brute force algorithm, it is still too time-consuming and intractable for our purpose. 
\subsection{Lower Bound}
In this subsection, we introduce an effective lower bound to reduce the number of times to compute the actual optimal Context-Aware Distance $d_{opt}(p,q)$. We would like to use the lower bound to determine the order of comparisons and hope to stop the search at the earliest time. We will explain this idea in detail in Section 4.3. 
Inspired by \cite{valmod}, instead of enumerating all possible combinations of context subsequences, we consider it as an optimization problem. To simplify the optimization process, we relax all the constraints and aim to find a pair of values $\mu'$ and $\sigma'$ which minimizes the value of $d_{opt}(p,q)$. Thus, the lower bound obtained from $\mu'$ and $\sigma'$ will be a global minimum value independent from data and smaller than the actual distance. The objective function is relaxed as shown as follows: 
\small\begin{equation}
\label{eqn:objective}
\scalebox{0.90}{$
\begin{split}
d_{opt}(p,q)& = \min\limits_{\Omega_{p,q}} d_{i,j}(p,q)\\
    & = \min\limits_{\mu_i,\mu_j,\sigma_i,\sigma_j}\sqrt{\sum_{m = 1}^l (\frac{t_{p+m-1} - \mu_i}{\sigma_i} - \frac{t_{q+m-1} - \mu_j}{\sigma_j})^2}\\
    & \geq \min\limits_{\mu', \sigma'}\bigg(\min\limits_{\mu_j, \sigma_j}\sqrt{\sum_{m = 1}^l (\frac{t_{p+m-1} - \mu'}{\sigma'} - \frac{t_{q+m-1} - \mu_j}{\sigma_j})^2}\bigg)\\
    & = \min\limits_{\mu_j, \sigma_j}\bigg(\min\limits_{\mu', \sigma'}D(\mu',\sigma',\mu_j,\sigma_j)\bigg) = LB(p,q),
\end{split}
$
}
\end{equation}    
where $LB(p,q)$ is the lower bound which we would like to achieve. 

We split the lower bound computation to two steps to optimize the equation. 
We first optimize the inner part $D^* = \min\limits_{\mu', \sigma'}D(\mu',\sigma',\mu_j,\sigma_j)$, and then solve the outer minimization problem $ \min\limits_{\mu_j, \sigma_j}D^*$ for $LB(p,q)$.
\subsubsection{Solving the Inner Problem}

To optimize the inner part $D^* = \min\limits_{\mu', \sigma'}D(\mu',\sigma',\mu_j,\sigma_j)$, we can simply get minimum value of $D(\mu',\sigma',\mu_j,\sigma_j)$ in terms of $\mu'$ and $\sigma'$ by solving $\frac{\partial D}{\partial \mu'} = 0$ and $\frac{\partial D}{\partial \sigma'} = 0$. We optimize on $D^2$ instead of $D$ for the simplicity of computation. Specifically, 
\begin{equation}
\label{eqn:D^2}
D^2  = \sum_{m = 1}^l (\frac{t_{p+m-1} - \mu'}{\sigma'} - \frac{t_{q+m-1} - \mu_j}{\sigma_j})^2.
\end{equation}
Starting by solving $\mu'$,
we have \begin{equation}
\label{eqn:mu}
\frac{\partial D^2}{\partial \mu'}  = \sum_{m=1}^{l} \frac{-2}{\sigma'} (\frac{t_{m+p-1}-\mu'}{\sigma'} - \frac{t_{m+q-1}-\mu_j}{\sigma_j}) = 0
\end{equation}
By solving equation (\ref{eqn:mu}), we can obtain $\mu^*$, the optimal $\mu'$ value, as:
\vspace{-0.5mm}
\begin{equation}\label{mu}
    \mu^* = \mu_p+\frac{\sigma'}{\sigma_j} (\mu_j - \mu_q).
\end{equation}
Substituting $\mu'$ with $\mu^*$ in equation~(\ref{eqn:D^2}), we get 
\begin{equation}
\label{eqn:D2}
\begin{split}
    D^2&= \sum_{m=1}^{l} (\frac{t_{p+m-1}-\mu_p}{\sigma'} - \frac{\mu_j - \mu_q}{\sigma_j}-\frac{t_{m+q-1}-\mu_j}{\sigma_j})^2\\
    &= \sum_{m=1}^{l} (\frac{t_{p+m-1}-\mu_p}{\sigma'} - \frac{t_{m+q-1}-\mu_q}{\sigma_j})^2.
\end{split}
\end{equation}
Intuitively, $\mu^*$ adjusts $T_p$ to the same offset of $T_q$ under context $C_j$ to minimize $D^2$. From equation \ref{eqn:D^2}, we see that $T_p$ and $T_q$ after simplification, have the same mean value. 

We next take derivative with regard to $\sigma'$ on equation \ref{eqn:D2} to solve $\frac{\partial{D(\mu',\sigma',\mu_j,\sigma_j)}}{\partial \sigma'} = 0$:
\small\begin{equation}
    \label{eqn:sigma'}
    \begin{split}
       \frac{\partial D^2}{\partial \sigma'} &= \frac{\partial}{\partial \sigma'}\sum_{m=1}^{l} (\frac{t_{p+m-1}-\mu_p}{\sigma'} - \frac{t_{m+q-1}-\mu_q}{\sigma_j})^2\\
       &=\frac{-2l\sigma_p}{(\sigma')^2} \bigg[\frac{\sigma_p}{\sigma'} - \frac{\sigma_q}{\sigma_j}\delta_{p,q} \bigg] = 0,
    \end{split}
\end{equation}
where $\delta_{p,q}$ is the correlation similarity between target subsequences $T_p$ and $T_q$: \begin{equation}
    \label{eqn:delta}
    \delta_{p,q} = \frac{\sum_{m=1}^{l}(t_{p+m-1}t_{q+m-1})- l\mu_{p}\mu_{q}}{l\sigma_p\sigma_q}
\end{equation}  By solving equation (\ref{eqn:sigma'}), we get $\sigma^*$, the optimal value of $\sigma'$ as:

\[ \sigma^* = \begin{cases} 
      \frac{\sigma_p\sigma_j}{\delta_{p,q}\sigma_q} & \delta_{p,q}> 0 \\
     \infty & \delta_{k,l}\leq 0
   \end{cases}
\]

Now we are ready to compute the square of the lower bound in equation (\ref{eqn:D2}) 
by the optimal $\mu^*$ and $\sigma^*$, \begin{equation}
\label{eqn:LB_tmp}
\begin{split}
    (D^*)^2 &= \overbrace{\frac{l\sigma^2_p}{(\sigma')^2} }^\text{D1}+  \overbrace{(-\frac{2l}{\sigma'\sigma_j}(\frac{QT_{p,q}}{l} - \mu_p\mu_q)))}^\text{D2} + \overbrace{\frac{l\sigma^2_q}{\sigma^2_j}}^\text{D3}\\
     & =  D1+D2+D3,
\end{split}
\end{equation}
where
\begin{equation*}
\label{eqn:lb_123}
\begin{split}
    D1 &= \frac{l\sigma^2_p}{(\sigma’)^2} = \frac{l\delta^2_{p,q}\sigma_q^2}{\sigma_j^2},\\
    D2 & = -\frac{2l\delta_{p,q}\sigma_q}{\sigma_p\sigma_j^2}\bigg(\frac{\frac{QT_{p,q}}{l} - \mu_p\mu_q))}{\sigma_p\sigma_q}\bigg) = -\frac{2l\delta^2_{p,q}\sigma_{q}^2}{\sigma_j^2}, \text{ and}\\
    D3 &= \frac{l\sigma_q^2}{\sigma_j^2}.
    \end{split}
\end{equation*}
Summing up $D1$, $D2$ and $D3$, we have
\begin{equation*}
    \begin{split}
    (D^*)^2    
    &=\frac{l\sigma_q^2(1-\delta^2_{p,q})}{\sigma_j^2}.
    \end{split}
\end{equation*}
After taking the square root, we obtain optimal $D^*$, \begin{equation}
\label{eqn:D*}
    D^* = \frac{\sigma_q}{\sigma_j}\sqrt{l(1-\delta^2_{p,q})}.
\end{equation}
Surprisingly, the optimal value of $D^*$ is very simple -- it only depends on the standard deviation of the reference context $\sigma_j$ and the reference target $\sigma_q$, rather than the mean values $\mu_j$ or $\mu_q$. This makes sense intuitively, as the ratio of standard deviations of target and its context reflects to some extent on the scale variant between the two targets under normalization of their respective contexts. 

\subsubsection{Solving the Outer Problem}
With the optimal value of $D^*$ for equation (\ref{eqn:objective}), we are now left with the outer problem $\min\limits_{\mu_j,\sigma_j}D^*(\mu_j, \sigma_j)$, which only depends on $\sigma_j$. We can obtain the minimum value by simply picking the maximum value of $\sigma_j$, that is $\frac{\sigma_q}{\max(\sigma_j)}\sqrt{l(1-\delta^2_{p,q})}$. 

Moreover, as our Optimal Context-Aware Euclidean Distance is symmetric, i.e. $d_{opt}(p,q) = d_{opt}(q,p)$, by symmetry, it is effortless to obtain the other version of lower bound, $\frac{\sigma_p}{\max(\sigma_i)}\sqrt{l(1-\delta^2_{q,p})}$. 
To obtain a tighter lower bound, we choose the larger $LB$ of the two versions. Therefore, our final lower bound $LB(p,q)$ is the following: 
\begin{equation}
\label{eqn: LB}
    LB(p,q) = \begin{cases} 
      \gamma_{i,j}(p,q)\sqrt{l(1-\delta^2_{p,q})} &  \quad\delta_{p,q}> 0, \\
     \gamma_{i,j}(p,q)\sqrt{l} & \quad\delta_{p,q} \leq 0.
   \end{cases}
\end{equation}
Where $\delta_{p,q}$ is the correlation as shown in equation (\ref{eqn:delta}), and $\gamma_{i,j}(p,q) = \max\bigg(\frac{\sigma_q}{\max(\sigma_j)}, \frac{\sigma_p}{\max(\sigma_i)}\bigg) $. We can compute $\gamma_{i,j}(p,q)$ with moving standard deviation in $O(1)$ complexity. Thus, we are able to compute the lower bound $LB(p,q)$ in $O(1)$, which is much less costly than computing the actual Optimal Context-Aware Distance as the latter requires $O(L^2)$ complexity. In the next section, we will utilize the lower bound $LB(p,q)$ to prune the unnecessary distance computations.  

\subsection{Proposed Pruning Algorithm}
Our algorithm is based on the proposed lower bound and Algorithm 1. For each target, we reorder all the reference targets according to their lower bound value in ascending order. If the target is compared with reference targets in this order, we can simply stop the search once the best-so-far distance is smaller than the current lower bound value, as the remaining unchecked actual distance must be greater than the best-so-far distance, hence significantly reduce the computation cost. 

\begin{algorithm}[!t]
    \caption{Proposed Algorithm}
  \begin{algorithmic}[1]
    \STATE \textbf{Input}: Time Series $\mathcal{T}$, Context Length $L$, Target Length $l$, $\Omega_{p,q}$
    \STATE \textbf{Output}: $best\_so\_far$, 
    $best\_match\_T$, $best\_match\_C$
    \STATE $best\_so\_far=0$
    \FOR{$p = 1$ to $|\mathcal{T}| - l +1$} 
    \STATE    $nn\_dist$ = $Inf$
        
        {\color{blue} /* Compute Lower  Bound based on equation (\ref{eqn: LB}) for target subsequences in $\mathcal{T}$*/}
    \STATE $LB=$ComputeAllLB($T_p$)
    \STATE $LB\_Sort$, $LB\_Index$=Sort(LB)
    \FOR{$k = 1$ to $|\mathcal{T}| - l +1$}
    \STATE $q=LB\_Index[k]$
     \IF{\textit{IsSelfMatch(p,q)}}  
     \STATE \textbf{continue}\quad {\color{blue} /* Skip Self-Matching */}
     \ENDIF
     \IF{$nn\_dist \leq LB\_Sort[k]$}
     \STATE \textbf{break}
     \ENDIF
     \FOR{$i = p-L+l$ to $p$} 
     \FOR{$j = q-L+l$ to $q$}
     \STATE $d_{i,j}(p,q)=$ $GetDist_{CA}$($p$, $q$, $i$, $j$, $\Omega_{p,q}$)
     \ENDFOR
     \ENDFOR
     
     \STATE $dist=\min_{i,j} d_{i,j}(p,q)$
     \IF{$dist < nn\_dist$}
     \STATE $nn\_dist=dist$
    \STATE $nn\_match\_T=(p,q)$ 
     \STATE $nn\_match\_C=(i,j)$
     \ENDIF
     \ENDFOR
     \ENDFOR
     \IF{ $nn\_dist > best\_so\_far$}
        \STATE $best\_so\_far = nn\_dist$
        \STATE $best\_match\_T = nn\_match\_T$
     \STATE $best\_match\_C = nn\_match\_C$
     \ENDIF

     \STATE \textbf{return} $best\_so\_far$, $best\_match\_T$, $best\_match\_C$
  \end{algorithmic}
\end{algorithm}

The proposed algorithm is shown in Algorithm 2. The largest closest match of the Optimal Context-Aware Distance is denoted as $best\_so\_far$ and is initialized to zero (Line 3). For every target subsequence $T_p$, the closest match distance of current target is initialized to infinity (Line 5). We then compute $LB(p,q)$ and sort all the reference targets in ascending order (Lines 6-7) with the cost of $O(n\log(n))$ and $T_q$ is checked in this order. Lines 8-28 are the inner loop which compares a target with all possible reference targets. Lines 10-12 check that $T_p$ and $T_q$ are from non-self match contexts to avoid trivial solution. Line 13 compares the $nn\_dist$ with the lower bound of the target $T_p$ with the current reference target $T_q$. If $nn\_dist$ is smaller or equal to the lower bound, it means that we have already found the closest match for the target $T_p$, so we can skip the rest of reference targets (Line 14) for current target $T_p$ and proceed to check the next target subsequence $T_{p+1}$; otherwise, the actual distance $dist$ is computed (Lines 16-21) and the $nn\_dist$ distance and the corresponding indices are updated accordingly (Lines 22-26). Finally, the global $best\_so\_far$ is updated based on $nn\_dist$ and is returned along with the indices of its corresponding contexts and targets (Lines 29-34). 

Overall, in the best case, for each target $T_p$, we only need to sort all lower bounds with the cost of $O(n(\log (n))$ and only compute actual distance once, which has a complexity of $O(L^2)$. Therefore, in the best case, the total cost for detecting semantic discord is $O(n^2\log(n))$, which is a great improvement compared with smart brute force $O(n^2L^2)$ as $\log n\ll L^2$ in most real world applications in time series\cite{yeh2016matrix}. We will show that the pruning rate of our algorithm reaches above 99.9\% in the experimental section.
\begin{figure}[h]
    \centering
    \includegraphics[width = 80mm]{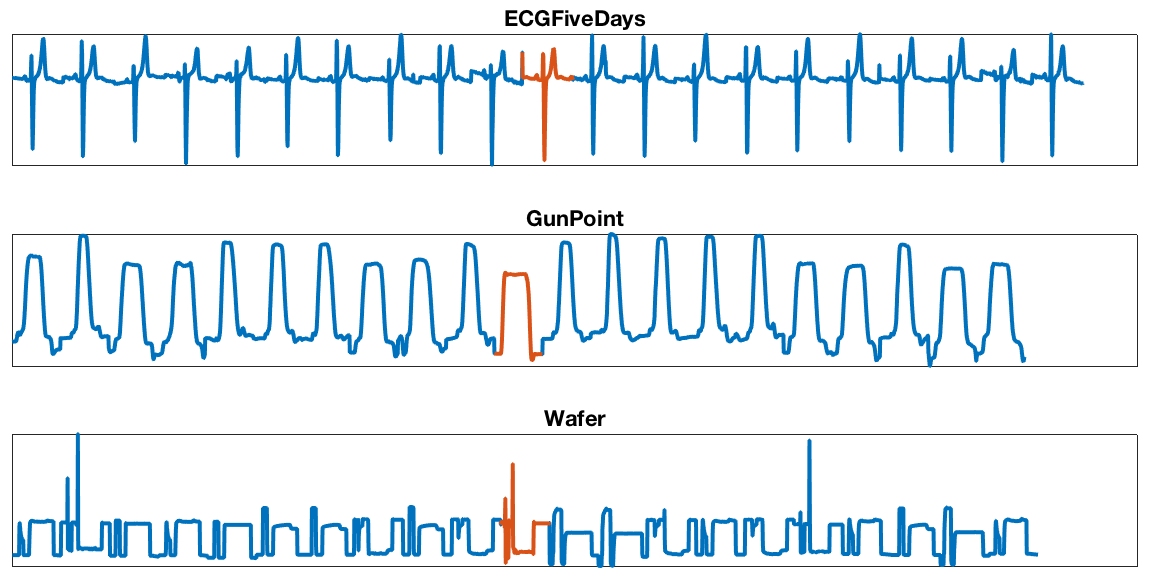}
    \caption{Time series generated from concatenation on different classes of UCR datasets. The subsequence in red is the ground truth of anomaly.}
    \label{fig:synthetic}
    \vspace{-2mm}
\end{figure}
\vspace{-2mm}
\section{EXPERIMENTS}
\subsection{Experiment Setup} We first describe the experiment setup on the proposed method and the baseline. Due to the unsupervised nature of discord, the ground truth is often not available. We use the widely-used UCR time series classification benchmark\footnote{https://www.cs.ucr.edu/$\sim$eamonn/time\_series\_data/} to generate time series. The datasets under the category of synthetic or frequency domain with high frequency noises and those with missing values are excluded as we are interested in real world data with interpretation in time domain. We select all the remaining data with lengths between 80 and 500. For each dataset, the time series is generated  by \textit{concatenating} twenty instances from one class and one instance from a different class. We randomly generate twenty time series for each dataset and report the average result. Figure \ref{fig:synthetic} shows three synthetic datasets that we generated from \textit{ECGFiveDays}, \textit{GunPoint} and \textit{Wafer} datasets.  
\begin{table*}[ht]
\caption{Datasets description and performance comparison}
\label{tab:main}
\resizebox{\textwidth}{!}{%
\begin{tabular}{@{}llllllllll@{}}
\toprule
Datasets                          & Type    & Length & Discord context & Discord target & RKNN context   & RKNN target & RD context     & RD target      & Proposed method \\ \midrule
Coffee                            & Spectro & 286    & 0.1             & 0.394          & 0.082          & 0.197       & 0              & 0.05           & \textbf{0.55}   \\
DistalPhalanxOutlineCorrect       & Image   & 80     & 0.283           & 0.234          & 0              & 0.05        & 0.05           & 0              & \textbf{0.3}    \\
ECG200                            & ECG     & 96     & 0.2             & 0.154          & 0.15           & 0.05        & 0.05           & 0              & \textbf{0.283}  \\
ECGFiveDays                       & ECG     & 136    & 0.172           & 0.4            & 0              & 0           & 0              & 0.066          & \textbf{0.65}   \\
GunPoint                          & Motion  & 150    & \textbf{0.461}  & 0.215          & 0              & 0.008       & 0.032          & 0              & 0.44            \\
GunPointMaleVersusFemale          & Motion  & 150    & 0.1             & 0.103          & 0              & 0           & 0              & 0.05           & \textbf{0.528}  \\
GunPointOldVersusYoung            & Motion  & 150    & 0.001           & 0.095          & 0              & 0.006       & 0.05           & 0.017          & \textbf{0.468}  \\
Ham                               & Spectro & 431    & \textbf{0.209}  & 0.05           & 0              & 0           & 0.034          & 0              & 0               \\
MiddlePhalanxOutlineCorrect       & Image   & 80     & 0.197           & 0.15           & 0              & 0.034       & 0.05           & \textbf{0.2}   & 0.198           \\
MoteStrain                        & Sensor  & 84     & 0.014           & \textbf{0.117} & 0.047          & 0           & 0              & 0              & 0.011           \\
PhalangesOutlinesCorrect          & Image   & 80     & 0.052           & \textbf{0.153} & 0              & 0           & 0              & 0.081          & 0.12            \\
PowerCons                         & Power   & 144    & 0               & 0.109          & 0.043          & 0           & 0              & 0              & \textbf{0.249}  \\
ProximalPhalanxOutlineCorrect     & Image   & 80     & \textbf{0.284}  & 0.05           & 0              & 0           & 0.05           & 0.203          & 0.059           \\
SonyAIBORobotSurface1             & Sensor  & 70     & 0               & 0.095          & \textbf{0.248} & 0.068       & 0.093          & 0.05           & 0.02            \\
SonyAIBORobotSurface2             & Sensor  & 65     & 0.017           & 0.05           & 0.019          & 0.05        & 0.05           & 0              & \textbf{0.194}  \\
Strawberry                        & Spectro & 235    & 0.341           & 0.418          & 0              & 0.02        & 0.009          & 0.093          & \textbf{0.452}  \\
ToeSegmentation1                  & Motion  & 277    & 0.008           & 0              & 0.077          & 0.029       & 0              & \textbf{0.188} & 0               \\
ToeSegmentation2                  & Motion  & 343    & 0               & \textbf{0.132} & 0.05           & 0.05        & 0.1            & 0.108          & 0.031           \\
TwoLeadECG                        & ECG     & 82     & 0.103           & 0.25           & 0              & 0.056       & 0              & 0.078          & \textbf{0.608}  \\
Wafer                             & Sensor  & 152    & 0.05            & 0.109          & 0              & 0.022       & 0.05           & 0.1            & \textbf{0.129}  \\
Wine                              & Spectro & 234    & \textbf{0.2}    & 0.1            & 0              & 0.039       & 0.1            & 0.116          & 0.053           \\
Yoga                              & Image   & 426    & 0.115           & 0.116          & 0              & 0           & \textbf{0.144} & 0              & 0.057           \\ \midrule
Number of Win                     &         &        & 4               & 3              & 1              & 3           & 1              & 2              & \textbf{11}     \\ \midrule
(Wilcoxon Test) Proposed vs Other &         &        & 0.0392          & 0.0386         & 0.0012         & 0.0003      & 0.0019         & 0.0078         &                 \\ \bottomrule
\end{tabular}%
}
\end{table*}
\vspace{-1.5mm}
\subsubsection{Baseline Methods} We compare the proposed approach with the following state-of-the-art discord detection methods : 
\begin{itemize}
    \item \textbf{HOTSAX}~\cite{keogh2005hot} identifies the anomaly by the largest 1-nearest-neighbor distance. Various papers \cite{zhumatrix,yeh2016matrix} have proposed acceleration methods on computing the 1-nearest-neighbor distances. We use \textit{STOMP}\cite{zhumatrix} as the accelerated version of HOTSAX to find the discord. We will use \textbf{Discord context} and \textbf{Discord target} to denote the lengths of discord context and target, respectively. 
    \item \textbf{RKNN}~\cite{lin2008exact} identifies the discord as the subsequence that has the lowest reverse nearest neighbor count. We use the exact version without approximating heuristic to achieve the best performance. The subsequence lengths of context and target, respectively, for the RKNN approach are denoted as \textbf{RKNN context} and \textbf{RKNN target}. 
    \item \textbf{Rule Density Curve}~\cite{senin2015time} uses Sequitur grammar rules density curve to identify the anomalous subsequences corresponding to rare grammar rules. We use \textbf{RD target} and \textbf{RD context} to denote the target length and context length, respectively, for the rule density curve method.  
\end{itemize}

The actual lengths of time series instances in the UCR datasets are used as the window sizes across all methods. Our proposed method uses the actual instance length as the context length, and we fix the target length arbitrary as 40\% of the context length across all datasets. The $\epsilon$ value is obtained by top 40\% percentile from 2,000 randomly-sampled contexts. 

To ensure a fair comparison, we run both the target and context lengths as window size for all baseline methods. For the additional parameters of the baselines, we follow the original papers and employ $k$ = 3 for RKNN, and use $w=4$ and $a=4$ for rule density curve.

\subsection{Performance measurement}
We use the overlapping rate of detected discord with the ground truth as the metric of evaluation. The overlapping rate is defined as $$Overlapping\ rate = \frac{|Discord \cap Ground Truth|}{|Discord|}.$$ The range is between 0 and 1, where a higher overlapping rate indicates a better performance. We pick overlapping rate as the evaluation criterion as it provides more direct quantitative measure on the quality of detection instead of simply setting a threshold on hit or miss. Moreover, overlapping rate allows effective measurement on anomaly length difference from the ground truth.     

The average overlapping rate of twenty synthetic time series generated from each dataset is reported. 

\subsection{Results}
Table~\ref{tab:main} shows the overlapping rate for all the methods on the generated datasets from the UCR database. Our proposed approach has the highest overlapping rate of 0.65 with the ground truth, and three results are higher than 0.5. In contrast, other methods mostly report overlapping rates that are lower than 0.4 with one exception of Discord Context on the GunPoint dataset. Discord with context length is the second best method which achieves 4 wins. Comparing with the three baseline methods, each with two sets of length options, Semantic Discord achieves the best overlapping rates on 11 datasets. From the result, we conclude that our method is better in terms of the overall wins, has the best performance compared to the state-of-the-art methods, and is able to identify anomalous subsequences that other methods could not find.  

We perform the Wilcoxon signed rank test on the result of overlapping rate. The $p$-value of Wilcoxon test between our proposed approach and all other methods are 0.0392, 0.00386, 0.0012, 0.03, 0.0019, 0.0078 respectively. All the $p$-values are less than 0.05 and we conclude that the proposed method significantly outperforms Discord, RKNN and Rule density curve. The results show that our proposed approach has statistically significant 'wins' over the other methods. 

Our method performs generally better on the datasets with some pattern structure with local differences such as ECG data, but it does not perform well in the cases where there is no existing semantic context information.

\subsubsection{Interpretability}
\begin{figure}[h]
    \centering
    \includegraphics[width=85mm]{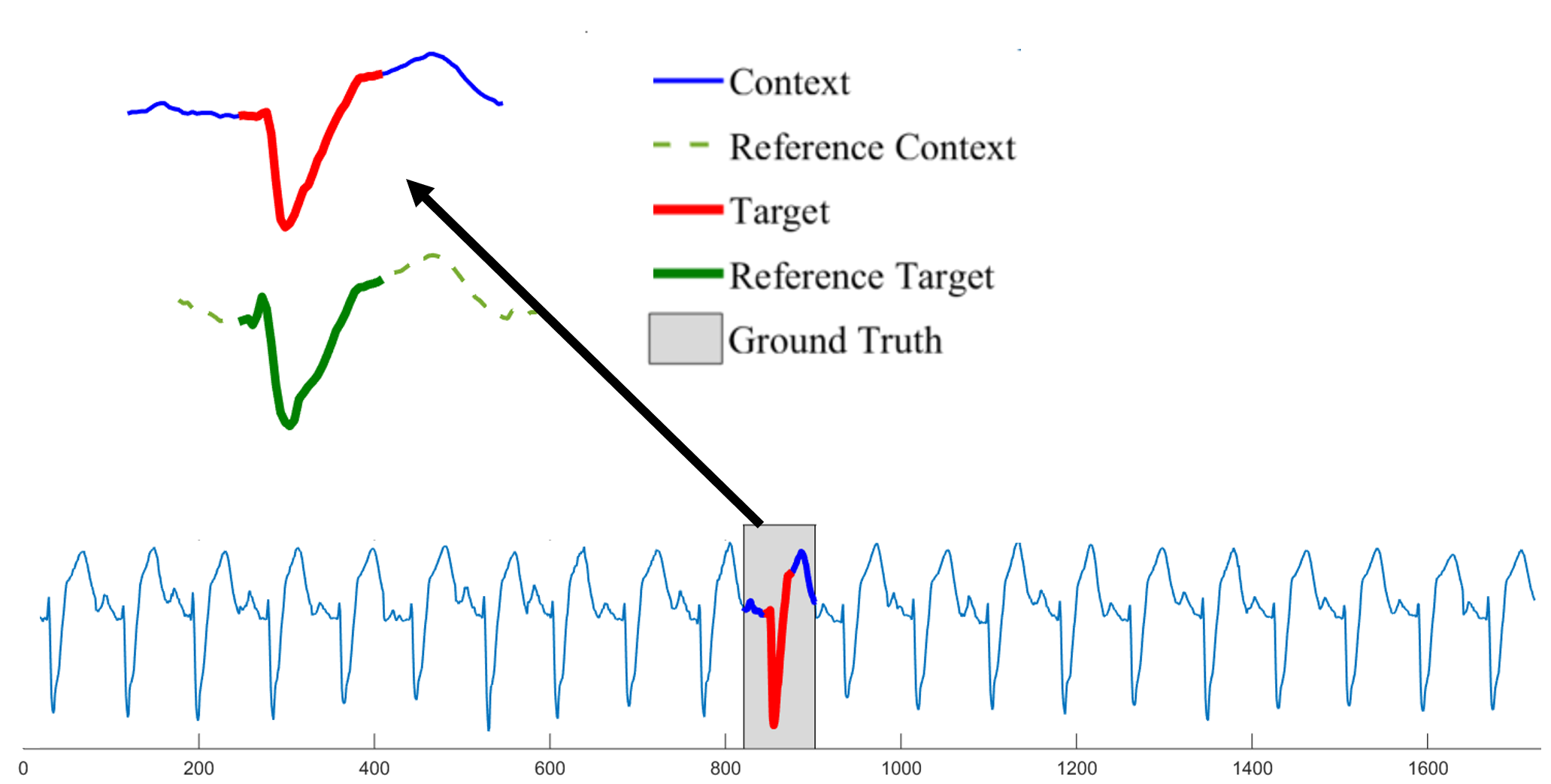}
    \caption{Target and context pair found agrees with ground truth on ECGFiveDays Synthetic data}
    \label{fig:ECG}
    \vspace{-1mm}
\end{figure}
Besides the superior performance, the Semantic Discord provides additional interpretability. Figure~\ref{fig:ECG} shows one of the synthetic data we generated from ECGFiveDays. The context subsequence successfully locates the anomaly at the exact location of the ground truth anomaly shown in the grey shaded area with starting position of 821. The normalized context and target pair are shown in red and blue, and the normalized reference target and reference context pair are drawn in light dotted green line and dark green line respectively. We can see that the beginning of the target subsequence in blue does not have the small sharp edge, and the ascend is straighter compared to the reference target. Our method provides insight to the analyst on the discovered semantic discord, as well as improved interpretability.
\vspace{-1mm}
\subsection{Scalability}
As we propose a new definition in this work, we compare the scalability against the smart brute force of our definition on both synthetic and real data. The data we use include Random Walk (a synthetic random walk data), ECG (an ECG trace of 530,000 samples), Dishwasher data (an electrical consumption data of 180,000 samples in Watts for 20 households at aggregate and appliance level \cite{murray2017electrical}). We test up to the first 128,000 samples and compare the total number of distance calls to the Euclidean distance function on the search for semantic discord. The length of the target discord is set to 160, and the context length is set to 400. All the experiments are performed on a computer with i5-7400HQ CPU of 2.80 Ghz with 32.0 GB memory on Matlab. 
\begin{figure}[h]
    \centering
    \includegraphics[scale=0.9, width = 85mm]{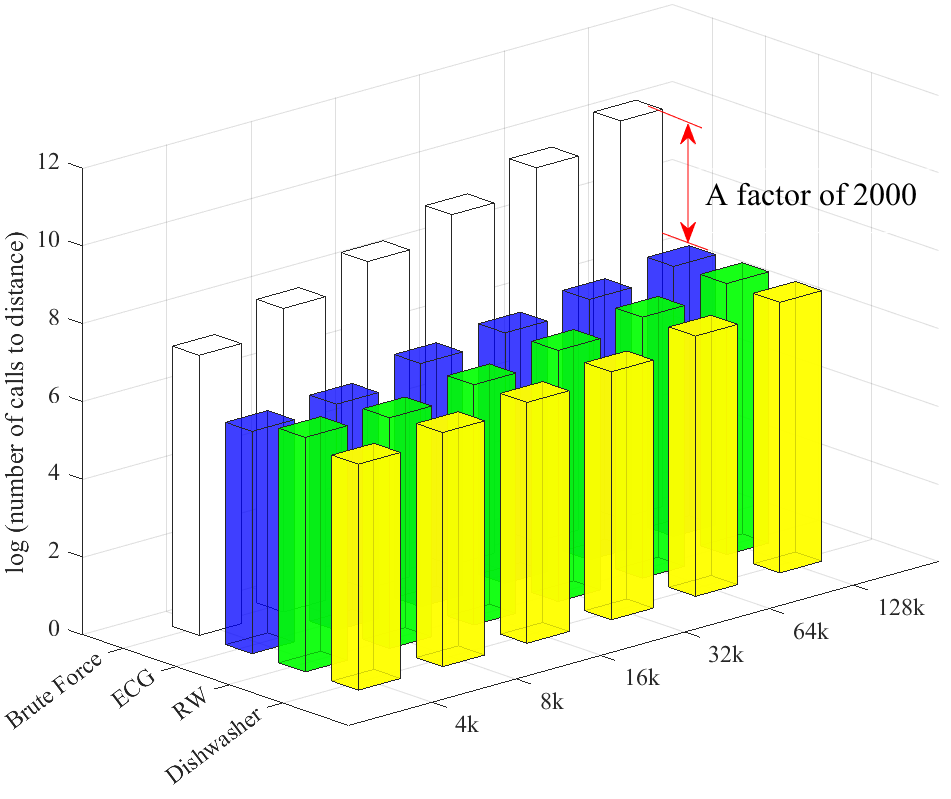}
    \vspace{-1mm}
    \caption{A comparison of our algorithm and brute force on number of distance calls (log scale) on three datasets. For brute force, the number of distance calls is the same for all datasets.}
    \label{fig:scalability}
    \vspace{-4mm}
\end{figure}

Figure \ref{fig:scalability} shows the numbers of distance calls by our method and the brute force algorithm, respectively. Under log scale, it is easy to observe that the difference increases as the dataset gets larger. At the length of 16,000, the pruning rate of our method is higher than 99.3\% for all three datasets. At the size of 128,000, our speed-up technique is about 2,000 times faster than brute force method for the ECG data, and the average pruning rate for the distance call is 99.95\%.

\section{CONCLUSION}
 Finding time series anomalous subsequence is a critical problem and has broad applications. In this work, we introduce a new definition named \textit{Semantic Discord}, which incorporates the context information from larger subsequences containing the anomaly candidates. We propose an efficient pruning algorithm with a derived lower bound that is up to 3 orders of magnitude faster compared to the smart brute force algorithm. Through the experiments, we demonstrate that our method outperforms the state-of-the-art methods, and is well suited for applications in different domains.
\vspace{-3mm}
\bibliography{reference}
\bibliographystyle{abbrv}
\end{document}